\documentclass{article}

\usepackage{arxiv}

\usepackage[utf8]{inputenc} 
\usepackage[T1]{fontenc}    
\usepackage{hyperref}       
\usepackage{url}            
\usepackage{booktabs}       
\usepackage{amsfonts}       
\usepackage{nicefrac}       
\usepackage{microtype}      
\usepackage{lipsum}
\usepackage{graphicx}
\usepackage{wrapfig}
\usepackage{enumitem}
\usepackage{appendix}
\usepackage{booktabs}
\usepackage{longtable}
\usepackage{authblk}
\usepackage{multirow}
\usepackage{xcolor}
\usepackage{algorithm2e}

\title{Mapping of Real World Problems to Nature Inspired Algorithm using Goal based Classification and TRIZ}

\author[1,2]{Palak Sukharamwala}
\author[1,3]{Manojkumar Parmar}
\affil[1]{Robert Bosch Engineering and Business Solutions Private Limited, Bangalore, India \\
\texttt{\{fixed-term.Sukharamwala.PalakPareshkumar, manojkumar.parmar\}@bosch.com}
}
\affil[2]{Institute of Technology, Nirma Univeristy, Ahmedabad\\ \texttt{18mcen16@nirmauni.ac.in}}

\affil[3]{HEC Paris, Jouy-en-Josas Cedex, France\\
\texttt{manojkumar.parmar@hec.edu}}

\begin{document}
\maketitle

\begin{abstract}
The technologies and algorithms are growing at an exponential rate. The technologies are capable enough to solve technically challenging and complex problems which seemed impossible task. However, the trending methods and approaches are facing multiple challenges on various fronts of data, algorithms, software, computational complexities, and energy efficiencies. Nature also faces similar challenges. Nature has solved those challenges and formulation of those are available as Nature Inspired Algorithms (NIA), which are derived based on the study of nature. A novel method based on TRIZ to map the real-world problems to nature problems is explained here.TRIZ is a Theory of inventive problem solving. Using the proposed framework, best NIA can be identified to solve the real-world problems. For this framework to work, a novel classification of NIA based on the end goal that nature is trying to achieve is devised. The application of the this framework along with examples is also discussed.
\end{abstract}

\newpage

\section{Introduction and Background}

Nature does things in an incredible way. Behind the visible phenomena, sometimes there are innumerable invisible causes. Scientists have been observing nature for hundred of years and trying to understand, explain, adapt and reproduce artificial systems based on it. There are countless living and non-living agents, act in parallel and sometimes against each other, to define nature and regulate the harmony. This is considered the dialectic of nature that resides in the concept of evolution of the natural world. The evolution of complexity in nature follows a distinctive order. There is also a distributed, self-organized and optimal processing of information in nature without any central control. The whole series of forms, mechanical, physical, chemical, biological and social, are distributed and aligned according to the complexity of the lowest to the highest. This sequence expresses its mutual dependence and its relation in terms of structure and history. Associated activities also change due to changing circumstances. All of these phenomena known or partially known so far emerge as new areas of study in  science and technology. Computer science helps to study nature-based problem solving techniques, underlying principles, mechanisms of natural, physical, chemical and biological organisms, who perform complex tasks appropriately with limited resources and capabilities. \cite{Siddique2015}

Science is a bridge between scientists and nature which has evolved over the centuries by enriching itself with new concepts, methods and tools and has developed into well-defined disciplines of scientific activity. Since then, humanity has been trying to understand nature by developing new tools and techniques. The field of nature-based computer science (NIC) is interdisciplinary in nature, combining computer science with knowledge from different branches of science, mathematics and engineering, which allows the development of new computational tools such as algorithms, hardware or software to solve problem. This chapter provides limitations of current technology, an overview of existing classification on Nature Inspired Algorithm (NIA), new approach called End Goal based classification, framework examples to understand use of framework.

\begin{table}[h]
    \centering
    \begin{tabular}{p{3cm} p{9cm}}
    \specialrule{.2em}{.1em}{.1em}
    Acronym & Full Name\\
    \specialrule{.1em}{.05em}{.05em}
    NIA & Nature Inspired Algorithm\\ \hline
    \multirow{2}{*}{TRIZ} & Russian : Teoriya Resheniya Izobretatelskikh Zadatch\\
        & English : Theory of inventive problem solving\\\hline
    AI & Artificial Intelligence\\\hline
    ML & Machine Learning\\\hline
    DL & Deep Learning\\\hline
    FOA & Fruit Fly Optimization\\\hline
    FOA-MHW & Fruit Fly Optimisation Algorithm - Multiplicative Holt-Winters\\\hline
    BA & Bat Algorithm\\\hline
    LSSVM & Least Square Support Vector Regression Model\\\hline
    MARS & Multivariate Adaptive Regression Splines\\\hline
    DP & Dynamic Programming \\\hline
    GA & Genetic Algorithm\\\hline
    TSP & Traveling Salesman Problem \\\hline
    ACO & Ant Colony Optimization \\
     \specialrule{.2em}{.1em}{.1em} 
    \end{tabular}
\caption{Acronyms used in chapter}
\end{table}

\section{Motivation behind NIA exploration}

In the first two decades of the 21st century, access to large amounts of data (known as "big data"), faster computers and advanced machine learning techniques were successfully applied to many problems for commercial benefits. The AI/ML algorithms and its applications are pervasive today, and they are solving many specific problems and making life easier. However, data scientist's favourite algorithms and many other technology from various engineering branches have their own limitations. We will discuss limitations in details in subsequent subsection.

\subsection{Prevailing Issues with technology}

\subsubsection{Data dependencies}

Data is the essence of AI/ML algorithm to achieve reasonable accuracy. Today's algorithms are highly data dependent. However, Issues like cost of acquisition of data, processing it, maintaining it and storing it in compliant way, makes it difficult to have sufficient amount of data many a times. The cost of data is one the biggest investment for any organization who want to leverage AI/ML. In absence of data, the accuracy of AI/ML algorithm suffers and renders them unfit for use \cite{redman_2018}. The key question is can we develop algorithms and alternatives which are not highly dependant on data, which can achieve "less data approach"? 

We need to be aware of the limitations of AI and where humans still need to take the lead. Data and algorithms cannot solve all type of problems. For a specific set of problems, the available set of algorithms fails to perform adequately despite of huge amount of available data \cite{lando_2018}. 

For a long time, Facebook believed that problems like the spread of misinformation and hate speech could be algorithmically identified and stopped. But under recent pressure from legislators, the company quickly pledged to replace its algorithms with an army of over 10,000 human reviewers. The medical profession has also recognised that AI cannot be considered a solution for all problems. The IBM Watson for Oncology program was a piece of AI that was meant to help doctors treat cancer. Even though it was developed to deliver the best recommendations, human experts found it difficult to trust the machine. As a result, the AI program was abandoned in most hospitals where it was trialled.

These examples demonstrate that there is no AI solution for everything. Not every problem is best addressed by applying machine intelligence to it. \cite{VYACHESLAV_2018}

\subsubsection{Demand of higher software complexity}
The increasing software demands also lead to increase in complexity.  The complexity is increasing at repaid rate as demand for intuitive and complex solutions is growing. The traditional methods of software programming and solutions are already proven inadequate to manage complexity \cite{braude2016software}. The question is can we develop software with reduced complexity yet rich in features?

\subsubsection{NP-Hard problems}
There are sets of intractable problems (NP-hard) which are not solvable due to computational complexity involved arising from known set of algorithms for example Traveling Salesman Problem. TSP is discussed in detail in following sections. The question is can we solve NP-Hard problems with available compute power today? \cite{vzerovnik2015heuristics}  \cite{woeginger2003exact}

\subsubsection{Energy Consumption}
Though DL algorithms are claimed to mimic human brains, still they lack in terms of efficiency and energy consumption. There is a long way to go to mimic human brains perfectly.

Despite the availability of data, algorithm and computational power, it is not possible to solve set of problems due to sheer amount of energy usage. The cost-benefit analysis produces unfavorable results due to impact of energy usage on environment \cite{hao_2019}. The key question is can we solve the problem in energy efficient way to match human brains energy efficiency frontier?

The discussed problems are considered small today; however, they are growing drastically. These problems indicate that growth of AI/ML based algorithms and applications are not sustainable from technical, business, and environmental point of view. Therefore to tackle the stated problem, there is a need to explore alternative solutions. Scientists believe NIA is the best suitable alternative.

\bigbreak

\subsection{Nature Inspired Algorithm at rescue}

Nature faces varied problems, and it has found the best way to solve them using constrained optimization over a time. The species on earth are doing various forms of optimization for rest of their life in their respective environment. Their survival is proof that the evolved optimizations are one of the best possible solutions. Nature has its way of transferring minimum intelligence from one generation to another using genes. Later on, life forms acquire higher amount of intelligence based on their experiences interacting with environment. In the entire process of acquiring intelligence, the usage of data is very minimum. At the same time, the decisions derived from the intelligence is good enough at the least. The life forms are capable of managing the complexity of real-world, and their decisions are achieved in the most efficient way as their survival depends on it.

As discussed, today’s digitally powered world is facing significant complex problems due to temporal and spatial complexity, variability, and constrained environments. The similarity between problems faced by nature and digital world is striking, and hence the proposition is to take over solutions of nature and implement it to solve digital problems. The solution formalized by studying nature is referred as Nature Inspired Algorithm (NIA) \cite{yang2010nature}. NIA are meta-heuristic algorithms, which provides approximate answers. They are designed to optimize numerical benchmark functions, multi-objective functions and solve NP-hard problems for a large number of variables, and dimensions.

The growth of current technologies is bound to diminish and will pave the way for new technologies to emerge. We believe that NIA is going to be the next disruptive technology to address the problems faced by current technologies and provide answers to key questions asked above.

\section{Novel TRIZ + NIA approach}

\subsection{Traditional Classification}

\begin{figure}
    \centering
    \centerline{\includegraphics[width=\textwidth]{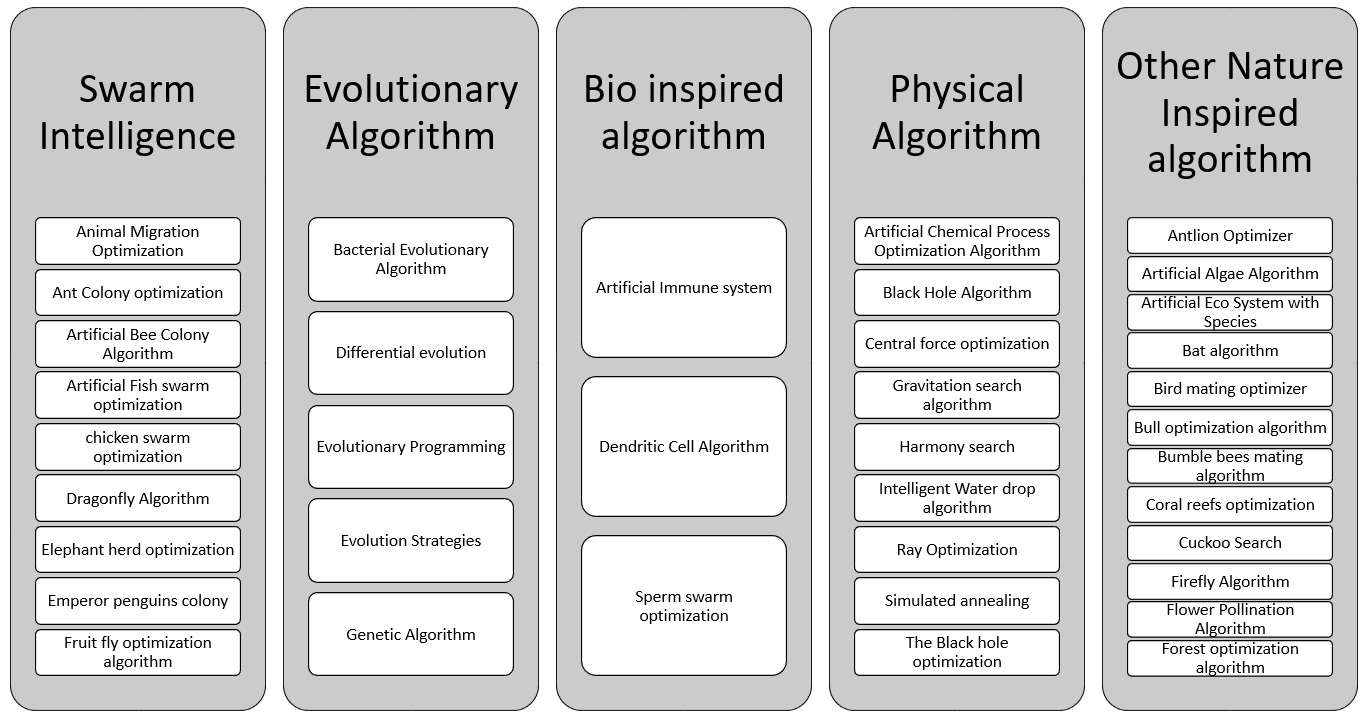}}
    \caption{Categories of Traditional Classification of NIAs}
    \label{fig:tradition_classificationl}
\end{figure}

There exist a classification for NIAs, which is solution based. It focuses on the techniques used by algorithms. According to traditional classification,Algorithms are classified into following classes.\cite{Siddique2015, nanda2014survey, binitha2012survey, fister2013brief}. Figure \ref{fig:tradition_classificationl} classifies NIA into below mentioned classes.

\begin{itemize}
    \item Swarm Intelligence
    \item Evolutionary Algorithms
    \item Bio-inspired Algorithms
    \item Physics-based Algorithm
    \item Other Nature Inspired Algorithms\\
\end{itemize}

\subsubsection{Swarm Intelligence}

Swarm intelligence is something in which agents work in parallel to achieve a certain task. Swarm Intelligence (SI) is simply the aggregate conduct of decentralized, self-organized entities. The similar idea is utilized for Artificial Intelligence in early days. Firstly it was presented by Gerardo Beni and Jing Wang in 1989, with regards to cellular robotic systems\cite{beni1993swarm}.

SI systems consist typically of a population of simple agents or boids interacting locally with one another and with their environment. The inspiration often comes from nature, especially biological systems. The agents follow simple rules, and although there is no centralized control structure dictating behaviour of individual agents.Their random interaction leads to emergence of "intelligent" global behavior, unknown to the individual agents. Examples of swarm intelligence in natural systems include ant colonies, bird flocking, hawks hunting, animal herding, bacterial growth, fish schooling and microbial intelligence.

The application of swarm principles to robots is called swarm robotics, while 'swarm intelligence' refers to the more general set of algorithms. 'Swarm prediction' has been used to solve forecasting problems. Similar approaches to those proposed for swarm robotics are considered for genetically modified organisms in synthetic collective intelligence\cite{sole2016synthetic}. Animal Migration Algorithm, Ant Colony Optimization, Artificial Fish Swarm Optimization, Fruit fly Optimization Algorithm, etc are example of Swarm Intelligence.

\subsubsection{Evolution Algorithm}

In artificial intelligence, an evolutionary algorithm (EA) is a subset of evolutionary computation,\cite{vikhar2016evolutionary}which represents a generic population-based meta heuristic optimization algorithm. An EA uses mechanisms inspired by biological evolution, such as reproduction, mutation, recombination, and selection. For optimization problem, candidate solution plays role of individual in population and the fitness function determines the quality of the solutions (see also loss function). Evolution of the population then takes place after the repeated application of the fitness function over generations.  

Evolutionary algorithms often perform well by approximating solutions to all types of problems as they ideally do not make any assumption about the underlying fitness landscape. Techniques from evolutionary algorithms applied to the modeling of biological evolution are generally limited to explorations of micro evolutionary processes and planning models based upon cellular processes. In most real applications of EAs, computational complexity is a prohibiting factor.\cite{cohoon2003evolutionary} In fact, this computational complexity is due to fitness function evaluation. Fitness approximation is one of the solutions to overcome this difficulty. However, seemingly simple EA can solve often complex problems \cite{cohoon2003evolutionary} like knapsack which is explained below. Therefore, there might not be any direct link between algorithm complexity and problem complexity. Bacterial Evolutionary Algorithm, Genetic Algorithm, etc. are famous Evolutionary Algorithm.

\subsubsection{Bio-inspired Algorithms}

Bio-inspired computing, short for biologically inspired computing, is a field of study which seeks to solve computer science problems using models of biology. It relates to social behavior, and emergence. Within computer science, bio-inspired computing relates to artificial intelligence and machine learning. Bio-inspired computing is a major subset of natural computation. In simpler words, Bio-inspired Algorithms imitates certain biological system of animal body. Artificial Immune System, Dendrite Cell system, etc. are example of Bio-Inspired Algorithms.

\subsubsection{Physics-based Algorithm}

Physics-inspired algorithms employ basic principles of physics, for example, Newton’s laws of gravitation, laws of motion and Coulomb’s force law of electrical charge. They are all based on deterministic physical principles. Black Hole Algorithm, Artificial Chemical Process Optimization Algorithm, Central force Optimization Algorithm, Gravitational Search Algorithm, etc. fall under this category.

\subsubsection{Other Nature Inspired Algorithms}

The set of algorithms which does not fit directly to above classification are put into this category. Artificial algae algorithm, Bat Algorithm,Coral Reef Optimization, Cuckoo Search, Firefly Algorithm, Flower Pollination Algorithm, etc. are popular example of Other Nature Inspired Algorithms.

\subsection{Limitation of traditional classification}

Drawback of the traditional classification is that it is not helpful for mapping of real life problem to conceptual problem. For an application, selection of algorithm is achieved using brute force. This classification does not make it easy to select algorithm.Hence the solution based approach is not ideal for mapping problem.

Drawback of tradition classifications are following :
\begin{itemize}
    \item It is solution based approach. The classification focuses on, how nature is solving an issue, not on what nature actually want to achieve.
    \item It is not helpful for mapping of real life problem to conceptual problem as it does not factor nature's problem.
\end{itemize}

\subsection{Combined approach NIA + TRIZ}

It is been analyzed that traditional classification and approaches lack a systematic method and framework to map real-world problems to NIA. They are mainly brute force in nature. Hence a novel approach to map real-world problems to NIA in a systematic way using TRIZ methodology \textcolor{blue}{is proposed}\cite{Palak2019mapping}. TRIZ principles are used along with new classification approach.

\subsubsection{TRIZ}

TRIZ is the Russian acronym for the "Theory of Inventive Problem Solving \cite{altshuller1999innovation}." TRIZ presents a systematic approach for understanding and defining challenging problems and it is the most useful in roles such as product development, design engineering, and process management. TRIZ includes a practical methodology, tool sets, a knowledge base, and model-based technology for generating innovative solutions for problem solving. 

It is useful for problem formulation, system analysis, failure analysis, and patterns of system evolution.The TRIZ was developed on three primary findings:

\begin{enumerate}
    \item Problems and solutions are repeated across industries and sciences
    \item Patterns of technical evolution are also repeated across industries and sciences
    \item The innovations generally use the solution and scientific finding from outside of focused field.\\
\end{enumerate}

\begin{figure}[h]
\centerline{\includegraphics[scale=0.50]{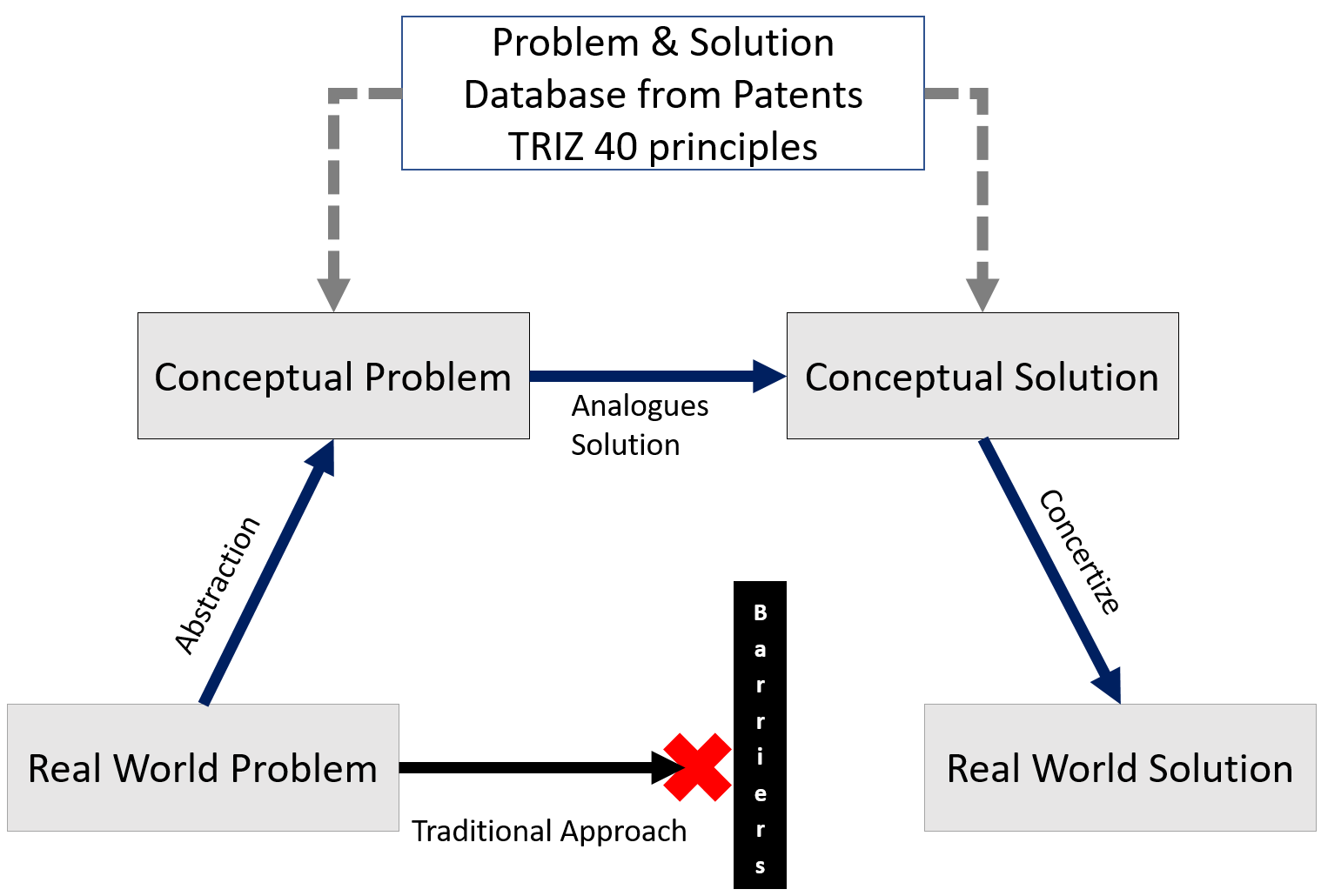}}
\caption{TRIZ Problem-Solution Approach \cite{nakagawa2005new}}
\label{Figure 4.1}
\end{figure}

Prism of TRIZ, as depicted in Figure \ref{Figure 4.1}. represents the 4 step approach for problem-solving. The real-world problem is mapped to conceptual problem using abstraction. The conceptual problem and corresponding solutions (40 TRIZ Principle) database is then used to find analogous solution to conceptual problem. The conceptual solution then is converted to a real-world solution. \textcolor{blue}{} In simple words, prizm TRIZ suggests to take help from solved problem to solve newer problems. Here, NIAs are used as solved solutions of nature’s problems.

\subsubsection{NIA + TRIZ}

\begin{figure}[h]
\centerline{\includegraphics[scale=0.4]{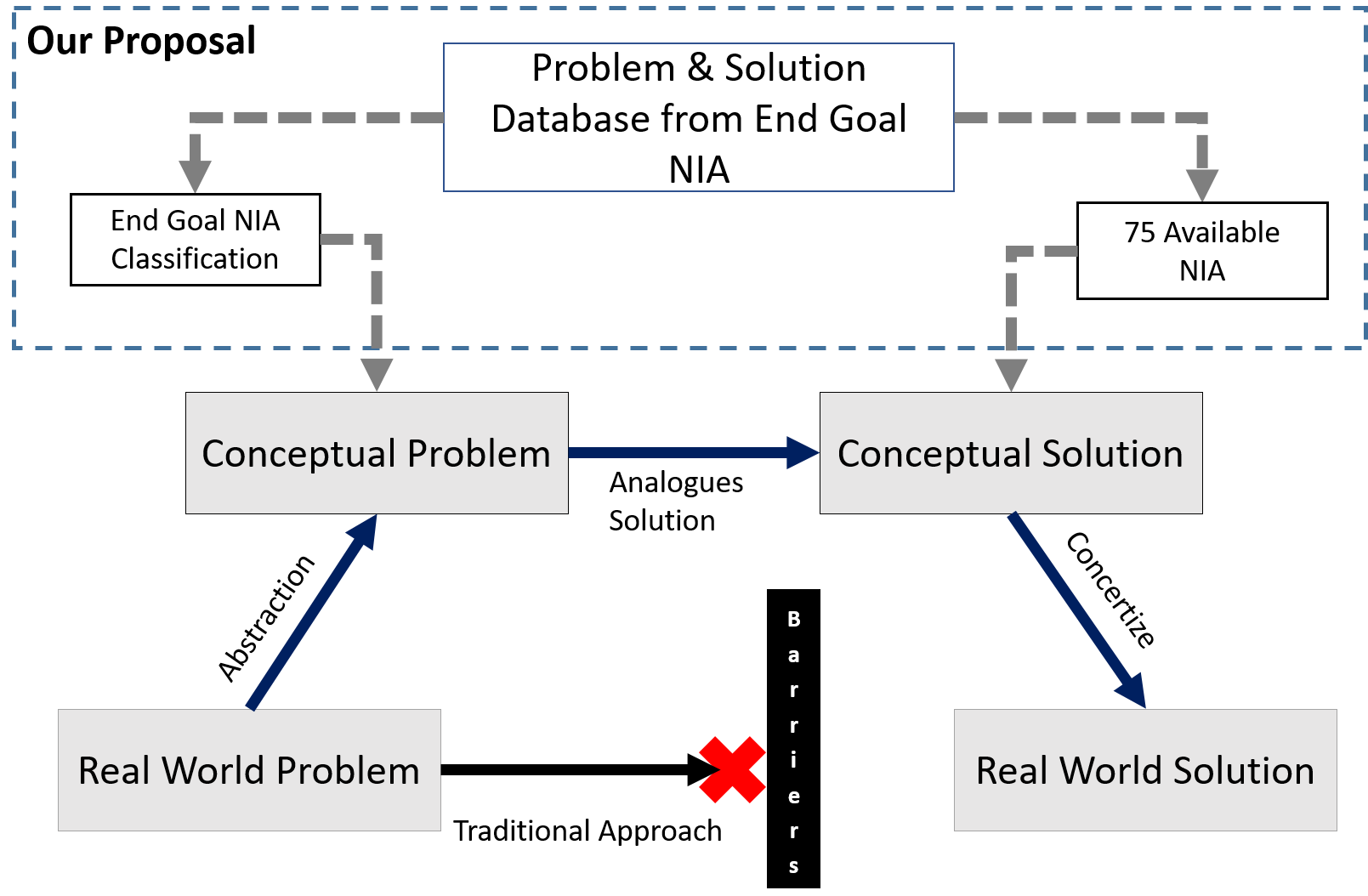}}
\caption{NIA+TRIZ Approach\cite{nakagawa2005new}}
\label{fig:Triz_Nia}
\end{figure}

\textcolor{blue}{Authors envisioned }that the TRIZ prism is the most suitable methodology to map real-world problems to nature problems and then provide corresponding solutions from NIA. A novel methodology as depicted in figure \ref{fig:Triz_Nia}, which combines TRIZ with end goal based classification. According to TRIZ foundation, if problems and solutions are repeated across industries and sciences, then existing pairs of problems and solutions can be used. For us that pair of problems and solutions are inspired from nature. However for this approach to work, we need new classification which is based on problems.For this reason, \textcolor{blue}{we} introduce novel classification. Here modified 4 step TRIZ process is explained.

\begin{enumerate}
    \item The real-world problem is mapped to conceptual problem using abstraction.
    \item The conceptual problem is then mapped to end goal of NIA.
    \item From an available database of NIA problem and solution; analogous NIA is derived into conceptual solution.
    \item The conceptual solution then is converted to a real-world solution.
\end{enumerate}

\subsection{End goal based Classification}

\begin{figure}[ht]\vspace*{4pt}
\centerline{\includegraphics[scale=0.65]{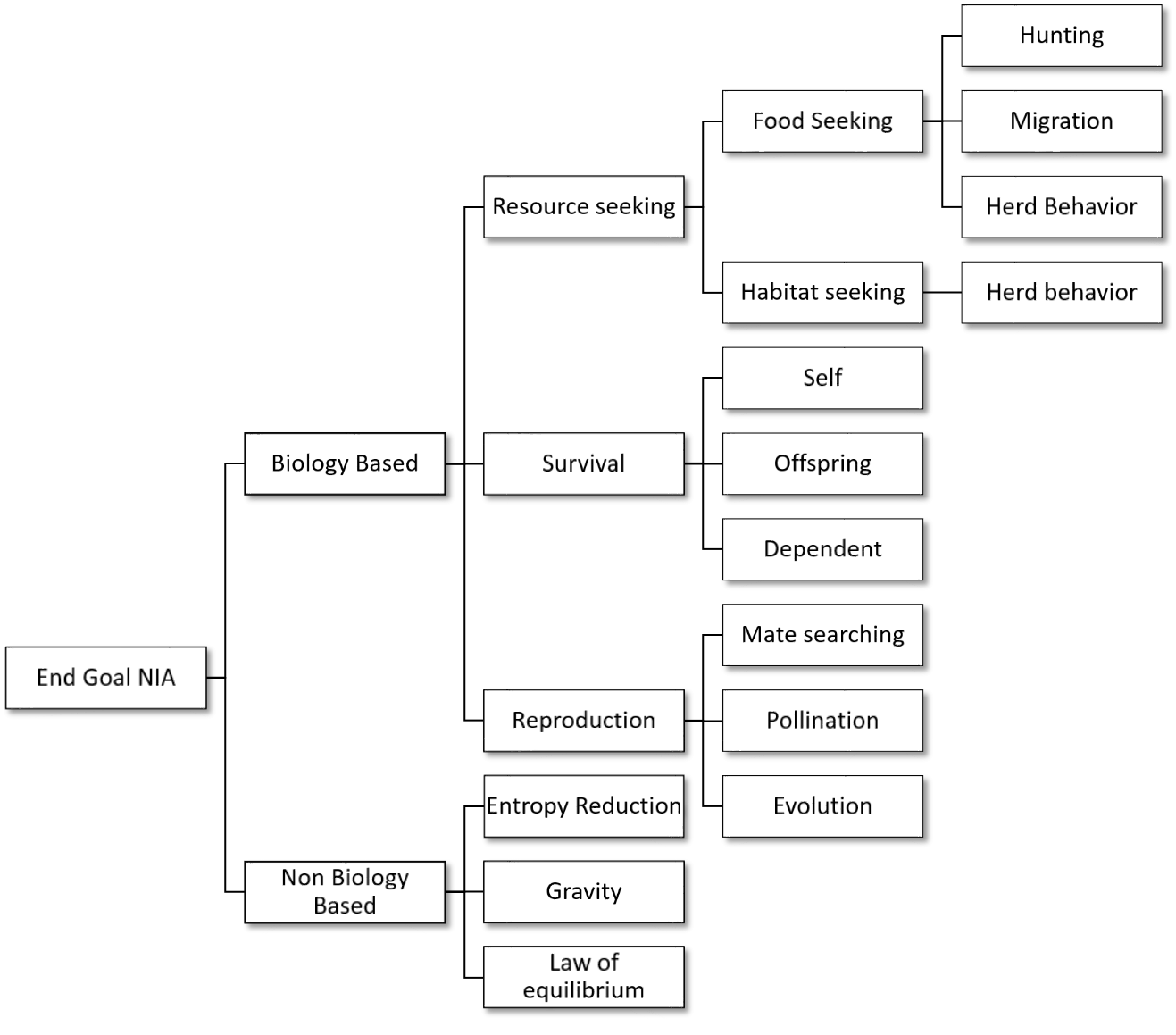}}
\caption{End goal based classification of NIA}
\label{fig:NIA_class}
\end{figure}

End goal based classification mainly focuses on problems nature has solved. It also considers the goal nature wants to achieve by solving the problem. The classification is 4 levels deep and varies based on goals and sub-goals, as depicted in Figure \ref{fig:NIA_class}. In total 75 NIA are classified using this approach and are present at one of the leaf nodes. Figure \ref{fig:NIA_class} represents classification diagram and Table \ref{Table(a)} \& \ref{Table(b)} explains the respective levels of classification.

\begin{enumerate}[label=Level\arabic*:,start=1]
    \item Biology and non biology based to distinguish living from non living
    \item Based on the primary goal
    \item Based on the sub goal
    \item Based on the behavior
\end{enumerate}

The detailed mapping of leaf nodes for NIA is available in Table \ref{Table(b)} for biology based and Table \ref{Table(a)} for non-biology based. For non-biology based, the classification is available based on primary goals only as sub goals and behavior has no real meaning.

\begin{table}[]
\caption{Biology based Algorithms}
\begin{tabular}{|l|l|l|l|}
\hline
\textbf{Level:2 Primary Goal}                       & \textbf{Level:3 Sub Goal}                         & \textbf{Level:4 Behavior}                        & \textbf{NIA}                                                \\ \hline
\multirow{22}{*}{Resource Seeking} & \multirow{19}{*}{Food Seeking}   & \multirow{6}{*}{Hunting}        & Antlion Optimizer                                  \\ \cline{4-4} 
                                   &                                  &                                 & Bat algorithm                                      \\ \cline{4-4} 
                                   &                                  &                                 & Grey wolf optimizer                                \\ \cline{4-4} 
                                   &                                  &                                 & Lion Optimization Algorithm                        \\ \cline{4-4} 
                                   &                                  &                                 & Salp swarm algorithm                               \\ \cline{4-4} 
                                   &                                  &                                 & Whale optimization algorithm                       \\ \cline{3-4} 
                                   &                                  & \multirow{2}{*}{Migration}      & Animal Migration Optimization                      \\ \cline{4-4} 
                                   &                                  &                                 & Artificial Algea Algorithm (AAA)                   \\ \cline{3-4} 
                                   &                                  & \multirow{11}{*}{Herd Behavior} & Ant Colony optimization                            \\ \cline{4-4} 
                                   &                                  &                                 & Artificial Bee Colony Algorithm                    \\ \cline{4-4} 
                                   &                                  &                                 & Artificial Fish swarm optimization                 \\ \cline{4-4} 
                                   &                                  &                                 & chicken swarm optimization                         \\ \cline{4-4} 
                                   &                                  &                                 & Dragonfly Algorithm                                \\ \cline{4-4} 
                                   &                                  &                                 & Fruit fly optimization algorithm                   \\ \cline{4-4} 
                                   &                                  &                                 & Gross hoper optimization                           \\ \cline{4-4} 
                                   &                                  &                                 & krill herd algorithm                               \\ \cline{4-4} 
                                   &                                  &                                 & Locust search algorithm                            \\ \cline{4-4} 
                                   &                                  &                                 & Particle swarm optimization algorithm              \\ \cline{4-4} 
                                   &                                  &                                 & Strawberry algorithm                               \\ \cline{2-4} 
                                   & \multirow{3}{*}{Habitat Seeking} & \multirow{3}{*}{Herd Behavior}  & Monarch butterfly optimization                     \\ \cline{4-4} 
                                   &                                  &                                 & Moth flame optimization algorithm                  \\ \cline{4-4} 
                                   &                                  &                                 & Sperm swarm optimization                           \\ \cline{1-4} 
\multirow{6}{*}{Survival}          & \multicolumn{2}{l|}{\multirow{3}{*}{Self}}                         & Artificial Immune system                           \\ \cline{4-4} 
                                   & \multicolumn{2}{l|}{}                                              & Dendritic Cell Algorithm                           \\ \cline{4-4} 
                                   & \multicolumn{2}{l|}{}                                              & Gross hoper optimization                           \\ \cline{2-4} 
                                   & \multicolumn{2}{l|}{\multirow{2}{*}{Offspring}}                    & Cuckoo Search                                      \\ \cline{4-4} 
                                   & \multicolumn{2}{l|}{}                                              & Emperor penguins colony                            \\ \cline{2-4} 
                                   & \multicolumn{2}{l|}{Dependant}                                     & Tree physiology optimization                       \\ \cline{1-4} 
\multirow{17}{*}{Reproduction}     & \multicolumn{2}{l|}{\multirow{4}{*}{Mating Searching}}             & Elephant herd optimization                         \\ \cline{4-4} 
                                   & \multicolumn{2}{l|}{}                                              & Firefly Algorithm                                  \\ \cline{4-4} 
                                   & \multicolumn{2}{l|}{}                                              & Honey bee mating optimization                      \\ \cline{4-4} 
                                   & \multicolumn{2}{l|}{}                                              & Social spider optimization                         \\ \cline{2-4} 
                                   & \multicolumn{2}{l|}{\multirow{11}{*}{Evolution}}                   & Artificial Eco System with Species                 \\ \cline{4-4} 
                                   & \multicolumn{2}{l|}{}                                              & Bacterial Evolutionary Algorithm                   \\ \cline{4-4} 
                                   & \multicolumn{2}{l|}{}                                              & Bird mating optimizer                              \\ \cline{4-4} 
                                   & \multicolumn{2}{l|}{}                                              & Bull optimization algorithm                        \\ \cline{4-4} 
                                   & \multicolumn{2}{l|}{}                                              & Bumble bees mating algorithm                       \\ \cline{4-4} 
                                   & \multicolumn{2}{l|}{}                                              & Coral reefs optimization                           \\ \cline{4-4} 
                                   & \multicolumn{2}{l|}{}                                              & Differencial evolution                             \\ \cline{4-4} 
                                   & \multicolumn{2}{l|}{}                                              & Evolutionary Programming                           \\ \cline{4-4} 
                                   & \multicolumn{2}{l|}{}                                              & Evolution Strategies                               \\ \cline{4-4} 
                                   & \multicolumn{2}{l|}{}                                              & Genetic Algorithm                                  \\ \cline{4-4} 
                                   & \multicolumn{2}{l|}{}                                              & Memetic algorithm                                  \\ \cline{2-4} 
                                   & \multicolumn{2}{l|}{\multirow{2}{*}{Pollinstion}}                  & Flower Pollination Algorithm                       \\ \cline{4-4} 
                                   & \multicolumn{2}{l|}{}                                              & Forest optimization algorithm                      \\ \hline

\end{tabular}
\label{Table(a)}
\end{table}

\begin{table}
\caption{Non-Biology based Algorithms}
\begin{tabular}{|p{0.3\linewidth}|p{0.65\linewidth}|}
\hline
\textbf{Level:2 Primary Goal}                        & \textbf{NIA}                                                \\ \hline
\multirow{3}{*}{Gravity}            & Black Hole Algorithm                               \\ \cline{2-2} 
                                    & Central force optimization                         \\ \cline{2-2} 
                                    & Gravitation search algorithm                       \\ \hline
\multirow{2}{*}{Entropy Reduction}  & Artificial Chemical Process Optimization Algorithm \\ \cline{2-2} 
                                    & Intelligent Water drop algorithm                   \\ \hline
\multirow{3}{*}{Law of Equilibrium} & Harmony search                                     \\ \cline{2-2} 
                                    & Water wave optimization                            \\ \cline{2-2} 
                                    & wind driven optimation                             \\ \hline
\end{tabular}
\label{Table(b)}
\end{table}

\section{Examples to support TRIZ+NIA approach}

\subsection{Fruit Optimization Algorithm to predict monthly electricity consumption }

In a paper, Jiang et al. \cite{jiang2019monthly} demonstrated use of fruit fly optimization algorithm to improve prediction of monthly electricity consumption with limited amount of training data. The proposed solution uses a hybrid forecasting model named FOA-MHW (Fruit Fly Optimisation Algorithm - Multiplicative Holt-Winters). The Holt-Winters algorithm is exponential smoothing algorithm for forecasting for time series data. The parameters of exponential smoothing are generated using FOA. 

The real-world problem of MHW is to find optimal parameters for smoothing with minimum amount of data. The parameter finding is converted into conceptual problem of Food-seeking problem which is under Resource seeking (Resource seeking -> Food-seeking problem -> Herd Behaviour) as referred in Figure \ref{fig:NIA_class}. For food-seeking in herd, the fruit fly is one of the superior species as it uses acute smell sensing in swarm with intelligent communication. Hence corresponding solution of fruit fly optimization is suitable for the identified problem. Therefore for the stated problem, FOA is found most suitable and outperformed traditional algorithms.

\begin{figure}[h]
\centerline{\includegraphics[scale=0.3]{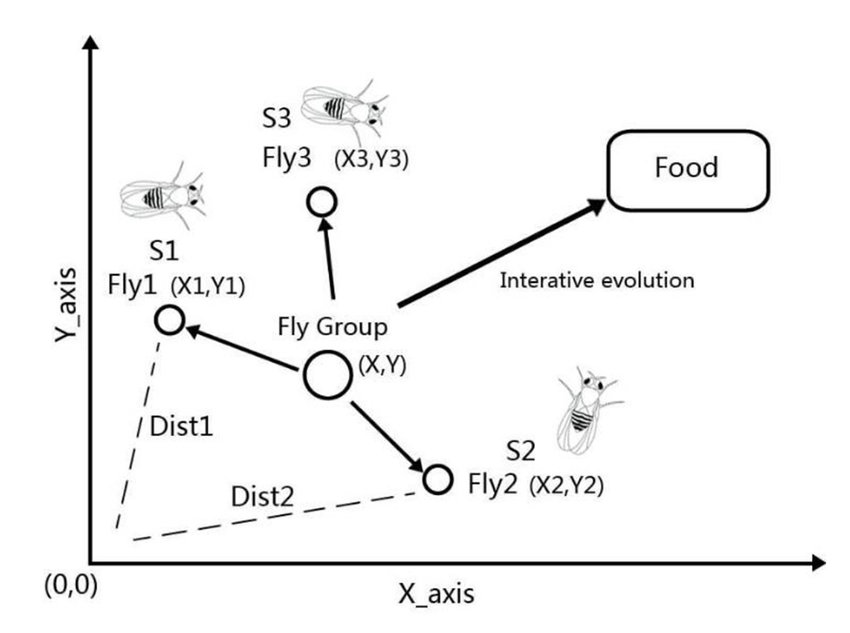}}
\caption{Diagram for Fruit fly optimization algorithm FOA}
\label{fig:FOA}
\end{figure}

\subsection{Bat Algorithm to model River Dissolved Oxygen Concentration}

Yaseen et al. \cite{yaseen2018integration} uses Bat Algorithm (BA) in Modeling River Dissolved Oxygen Concentration. Here NIA is integrated with Least Square Support Vector Regression Model. The accuracy of LSSVM-BA model compared with those M5 tree and MARS models are found  to increase by 20\% and 42\%, respectively, in terms of root-mean-square error. 

Studies have reported that the efficiency of LSSVM models significantly depends on the values of the kernel and regularization parameters. The hyper-parameters of LSSVM can be considered as decision variables and should be determined accurately by optimization algorithms for better performance of LSSVM models. In this study, the hyper-parameters of LSSVM were optimized using the BA. In BA, Bat's hunting behaviour is mimicked. Bat hunting comes under resource seeking followed by food seeking category(Resource seeking -> Food-seeking problem -> Hunting) as referred in Figure \ref{fig:NIA_class}, So it can be concluded that parameter optimization problem can be mapped with food seeking problem.

\begin{figure}[h]
\centerline{\includegraphics[scale=0.50]{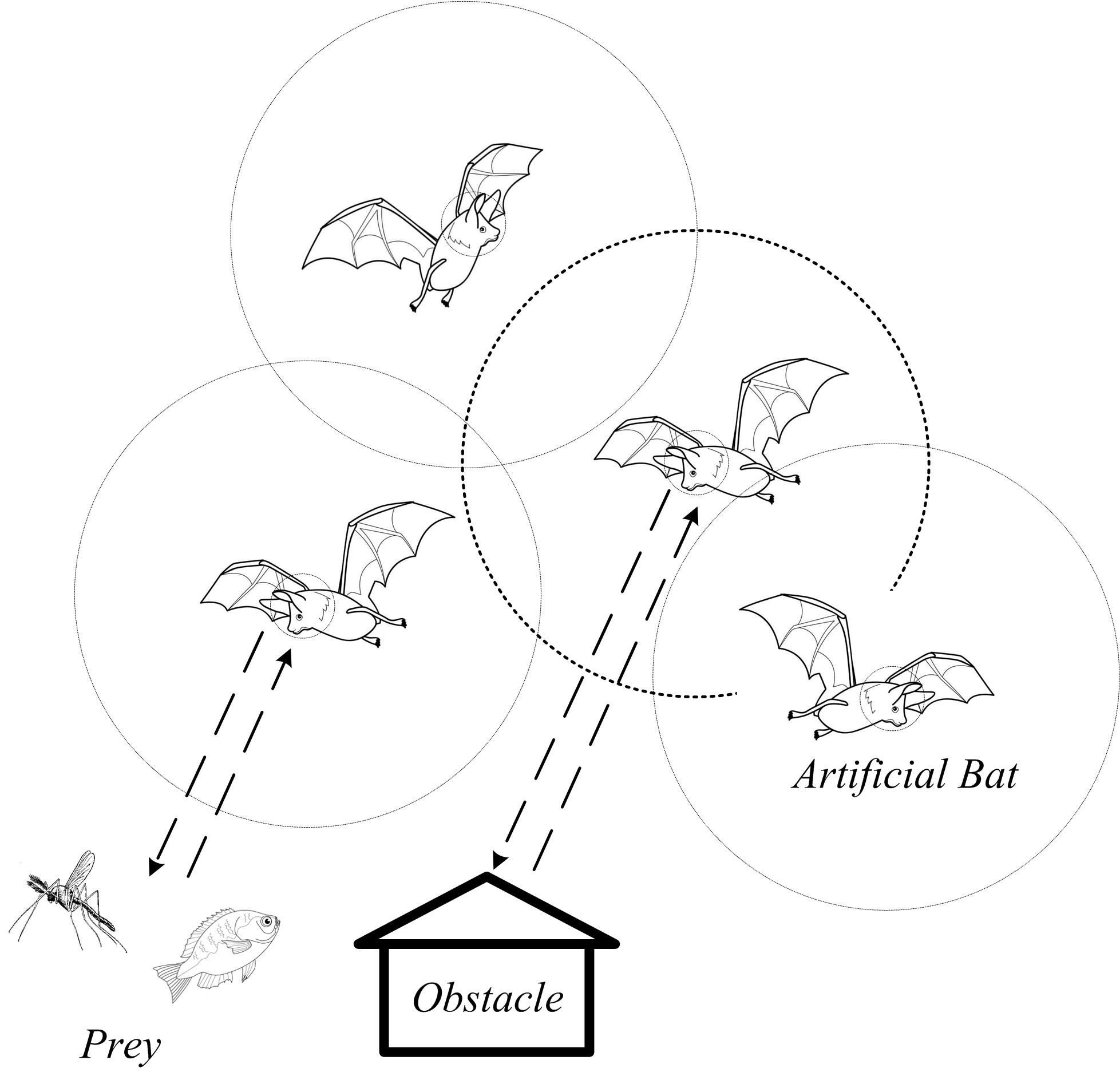}}
\caption{Diagram for Bat Algorithm}
\label{fig:BA}
\end{figure}

\subsection{Genetic Algorithm to tune the structure and parameters of a neural network}

Frank et al.\cite{leung2003tuning} discuss tuning  of Neural Network parameters using Genetic Algorithm. This is the very first paper from 2003 where it was proposed to use NIA to train network.

Neural network is proved to be a universal approximator. A three-layer feed forward neural network can approximate any nonlinear continuous function to an arbitrary accuracy. However, a fixed structure may not provide the optimal performance within a given training period. A small network may not provide good performance owing to its limited information processing power. A large network, on the other hand, may have some of its connections redundant. Moreover, the implementation cost for a large network is high. It could be the best if algorithm suggests the best structure for a neural network.It can lead to low cost of implementing the neural network, in terms of hardware and processing time.

\begin{figure}[h]
\centerline{\includegraphics[scale=0.85]{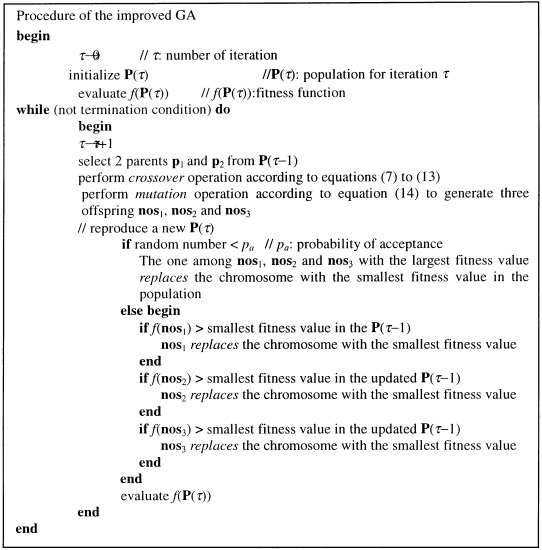}}
\caption{Procedure of improved Genetic Algorithm}
\label{fig:improved_GA}
\end{figure}

Here parameters like number of neurons, number of level, dense layer activation function and network optimizer are presented in one array form. Array is also used to present output solution. Choosing the correct representation of an output solution is very important in Nature Inspired Algorithms. In initialization, any random value for these parameter is taken. Priory can also be used instead of random values. Networks are trained using these parameter. The difference between predicted and actual value is fitness function. The change values of parameter is according to improved Genetic Algorithm. Figure \ref{fig:improved_GA} depicts pseudo algorithm along with procedure.

If we talk about abstraction from real world problem to Nature issues, then we can say the best structure is the result of survival among rest. self survival can be mapped with survival of the best structure. That is the reason GA algorithm from Biology based -> survival ->self category as referred in Figure \ref{fig:NIA_class} is chosen.

\section{Solution of NP-H using NIA}

\subsection{0-1 Knapsack Problem}

The knapsack problem is part of Combinatorial optimization problems' family. Here, 0-1 knapsack problem is one of the variant of knapsack problem. Knapsack problems appear in real-world decision-making processes in a wide variety of fields. Few traditional applications are finding the least wasteful way to cut raw materials, in the construction, scoring of tests in which the test-takers have a choice as to which questions they answer, etc.\cite{KPwiki} The knapsack problem has been studied for more than a century. Computer scientists always has fascination for knapsack problem because the decision problem form of the knapsack problem is NP-complete, thus there is no known algorithm both correct and fast (polynomial-time) in all cases.

The knapsack problem is defined as a set of items(xi) is given, each with a weight(wi) and a value(vi). Determine the number of each item to include in a collection so that the total weight is less than or equal to a given capacity(W) and the total value is as large as possible. in 0-1 knapsack the condition is that each and every chosen items must be whole, fraction of an item cannot be selected in solution.\cite{KPwiki}

Dynamic programming solution for the 0-1 knapsack problem also runs in pseudo-polynomial time. Solution runs in $O(nW)$ time and $O(nW)$ space. Another algorithm for 0-1 knapsack, discovered in 1974 and sometimes called "meet-in-the-middle" due to parallels to a similarly named algorithm in cryptography, is exponential in the number of different items but may be preferable to the DP algorithm when capacity$(W)$ is large compared to number of total itmes$(n)$. The algorithm takes O$(2^{n/2})$ space and O$(n2^{n/2})$ time. George Dantzig proposed a greedy approximation algorithm to solve the knapsack problem, but for the bounded problem, where the supply of each kind of item is limited, the algorithm is incapable to give optimal solution.\cite{KPwiki} These results concludes thatDynamicc Programming is the best approach among the traditional algorithms to treat the knapsack problem.

\begin{figure}[ht]
\centerline{\includegraphics[scale=0.9]{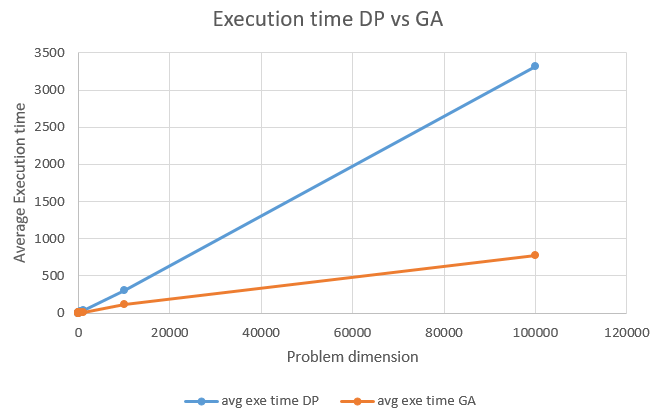}}
\caption{Execution time comparison between Dynamic Programming and Genetic Algorithm}
\label{fig:GAvsDP}
\end{figure}

\begin{figure}[h]
\centerline{\includegraphics[scale=0.42]{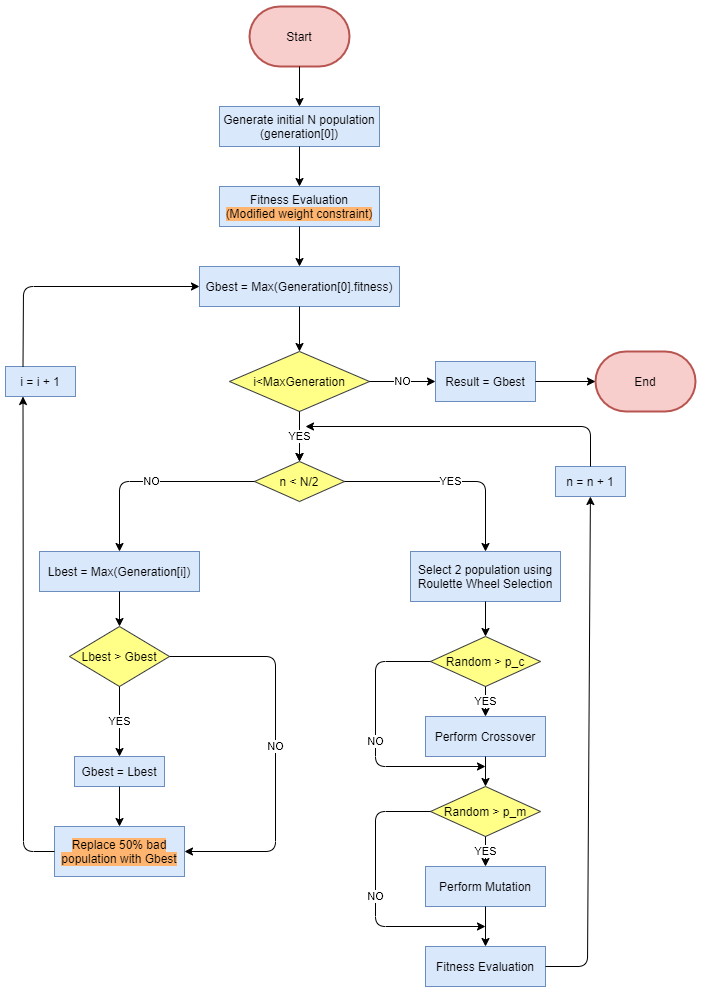}}
\caption{Flow chart of Genetic Algorithm to solve 0-1 knapsack problem}
\label{fig:GA_KP}
\end{figure}

Generally knapsack is an packing kind of problem. As we all know some operations gets better over time. Experience makes those task to perform better and to take better decisions. Experience comes with the the generations and generations are part of evolution. So here we can take help of Biology based algorithms -> Reproduction -> Evolutionary algorithms for knapsack problem. In which genetic algorithm is one of the best choice. Figure \ref{fig:GA_KP} shows flow diagram to use Genetic Algorithm to solve 0-1 knapsack problem.

When this flow diagram is implemented. It shows better results then Dynamic programming. Figure \ref{fig:GAvsDP} shows the execution time compression. Here, problem size is the total number item, so it is observable that slope of execution time for GA is is lesser than slope of execution time for DP. In terms of execution time GA performs much better than DP.

\subsection{Travelling Salesman Problem}

Travelling Salesman is also an NP-hard problem in combinatorial optimization, important in operations research and theoretical computer science. In the theory of computational complexity, the decision version of the TSP (where given a length L, the task is to decide whether the graph has a tour of at most L) belongs to the class of NP-complete problems. Thus, it is possible that the worst-case running time for any algorithm for the TSP increases superpolynomially (but no more than exponentially) with the number of cities. The TSP has several applications even in its purest formulation, such as planning, logistics.

TSP is defined as a list of cities and the distances between each pair of cities is given and the question is to find the shortest possible route that visits each city and returns to the origin city.

The most direct solution would be to try brute force approach. Testing of all permutations (ordered combinations) takes running time of a polynomial factor of $O(n!)$, the factorial of the number of cities, so this solution becomes impractical even for only 20 cities. Another approach is branch-and-bound algorithms, which can be used to process TSPs containing 40–60 cities.

Artificial intelligence researcher Marco Dorigo described in 1993 a method of heuristically generating "good solutions" to the TSP. As we all know NIAs are the best meta-heuristic algorithms. Route finding task is matter of team work, simultaneous searching helps to achieve the best solution. Herd behaviour can be helpful for this type of problem. Scientists' concerned problem is already solved by nature. Ants are the best solver of route optimization problem. Hence (Biology-based algorithms -> Resource seeking -> Herd Behaviour -> Ant Colony Optimization) leads to the end of our algorithm search. 

Ant Colony Optimization Algorithm sends out a large number of virtual ant agents to explore many possible routes on the map. Each ant probabilistically chooses the next city to visit based on a heuristic combining the distance to the city and the amount of virtual pheromone deposited on the edge to the city. The ants explore, depositing pheromone on each edge that they cross, until they have all completed a tour. At this point the ant which completed the shortest tour deposits virtual pheromone along its complete tour route (global trail updating). The amount of pheromone deposited is inversely proportional to the tour length: the shorter the tour, the more it deposits.

\textcolor{blue}{Manhattan distance and Euclidean Distance}

\section{Conclusion}

The sole purpose of introducing 'The End Goal based Classification' is to convey the missing pattern to link problem and solution. The identification of the problem and classification of problem helps to narrow down the range of suitable NIAs. A task like Parameter tuning, structure selection, parameter optimization which follows brute force approach can be solved with the help of NIA classification and mapping with conceptual problems. This approach is expected to benefit researchers and engineers working on computationally intensive and data starved problems to identify solutions in the most efficient way.

\bibliography{references}
\bibliographystyle{unsrt}

\end{document}